\documentclass[sigconf]{acmart}
\usepackage{tabularx}
\usepackage{multicol}
\usepackage{multirow}
\usepackage{booktabs}
\usepackage{threeparttable}

\usepackage{algorithmicx,algorithm}
\usepackage[utf8]{inputenc}
\usepackage[noend]{algpseudocode}
\usepackage{algorithmicx,algorithm}

\usepackage{url}
\usepackage{hyperref}
\usepackage{verbatim}
\usepackage{tabularx}
\usepackage{lineno}
\usepackage{amsfonts}
\usepackage{graphicx}
\usepackage{amsmath}
\usepackage[noend]{algpseudocode}
\graphicspath{ {./images/} }
\AtBeginDocument{%
  \providecommand\BibTeX{{%
    \normalfont B\kern-0.5em{\scshape i\kern-0.25em b}\kern-0.8em\TeX}}}

\copyrightyear{2021}
\acmYear{2021}
\setcopyright{acmcopyright}\acmConference[ICMI '21]{Proceedings of the 2021 International Conference on Multimodal Interaction}{October 18--22, 2021}{Montréal, QC, Canada}
\acmBooktitle{Proceedings of the 2021 International Conference on Multimodal Interaction (ICMI '21), October 18--22, 2021, Montréal, QC, Canada}
\acmPrice{15.00}
\acmDOI{10.1145/3462244.3479931}
\acmISBN{978-1-4503-8481-0/21/10}

\begin{document}

\title{Graph Capsule Aggregation for Unaligned Multimodal Sequences}

\author{Jianfeng Wu}
\email{wujf36@mail2.sysu.edu.cn}
\affiliation{%
  \institution{Sun Yat-sen University}
  %\streetaddress{P.O. Box 1212}
  \city{Guangzhou}
  \state{Guangdong}
  \country{China}
 }

\author{Sijie Mai}
\email{maisj@mail2.sysu.edu.cn}
\affiliation{%
  \institution{Sun Yat-sen University}
  %\streetaddress{P.O. Box 1212}
  \city{Guangzhou}
  \state{Guangdong}
  \country{China}
 }

 \author{Haifeng Hu}
 \authornote{Haifeng Hu is the corresponding author.}
\email{huhaif@mail.sysu.edu.cn}
\affiliation{%
  \institution{Sun Yat-sen University}
  %\streetaddress{P.O. Box 1212}
  \city{Guangzhou}
  \state{Guangdong}
  \country{China}
 }


\renewcommand{\shortauthors}{Trovato and Tobin, et al.}

\begin{abstract}
Humans express their opinions and emotions through multiple modalities which mainly consist of textual, acoustic and visual modalities. Prior works on multimodal sentiment analysis mostly apply Recurrent Neural Network (RNN) to model aligned multimodal sequences. However, it is unpractical to align multimodal sequences due to different sample rates for different modalities. Moreover, RNN is prone to the issues of gradient vanishing or exploding and it has limited capacity of learning long-range dependency which is the major obstacle to model unaligned multimodal sequences. In this paper, we introduce \textbf{\underline{Graph}} \textbf{\underline{C}}apsule \textbf{\underline{Ag}}gr\textbf{\underline{e}}gation (GraphCAGE) to model unaligned multimodal sequences with graph-based neural model and Capsule Network. By converting sequence data into graph, the previously mentioned problems of RNN are avoided. In addition, the aggregation capability of Capsule Network and the graph-based structure enable our model to be interpretable and better solve the problem of long-range dependency. Experimental results suggest that GraphCAGE achieves state-of-the-art performance on two benchmark datasets with representations refined by Capsule Network and interpretation provided.
\end{abstract}


\begin{CCSXML}
<ccs2012>
<concept>
<concept_id>10002951.10003227.10003251.10003255</concept_id>
<concept_desc>Information systems~Multimedia streaming</concept_desc>
<concept_significance>500</concept_significance>
</concept>
</ccs2012>
\end{CCSXML}

\ccsdesc[500]{Information systems~Multimedia streaming}

\keywords{Multimodal Sentiment Analysis; Unaligned Multimodal Sequences; Graph Capsule Aggregation; Long-range Dependency}

%% \begin{teaserfigure}
%%   \includegraphics[width=\textwidth]{sampleteaser}
%%   \caption{Seattle Mariners at Spring Training, 2010.}
%%   \Description{Enjoying the baseball game from the third-base
%%   seats. Ichiro Suzuki preparing to bat.}
%%   \label{fig:teaser}
%% \end{teaserfigure}

\maketitle

\section{Introduction}
Humans analyze sentiment by the rich information from spoken words, facial attributes and tone of voice, which correspond to textual, visual and acoustic modalities, respectively\cite{1994Tools,2014The}. It is natural that multimodal sources provide more reliable information for a model to predict sentiment labels. Nevertheless, there are two fundamental challenges for multimodal sentiment analysis. One is the “unaligned” nature of multimodal sequences. For instance, streams from audio and vision are created by receptors using different receiving frequency. As a result, successfully inferring long-range dependency is the key to tackle the issue of “unaligned” nature. The other challenge is how to effectively and efficiently model the long  sequences. As common methods to model sequences, RNN and its variants are susceptible to gradient vanishing or exploding and have high time complexity due to their recurrent nature \cite{DBLP:journals/corr/abs-2011-13572}. Therefore, it is critical to propose a model which can process sequential data appropriately without recurrent architecture.
 \begin{figure}[t]
    \centering
    \includegraphics[width=0.5\textwidth]{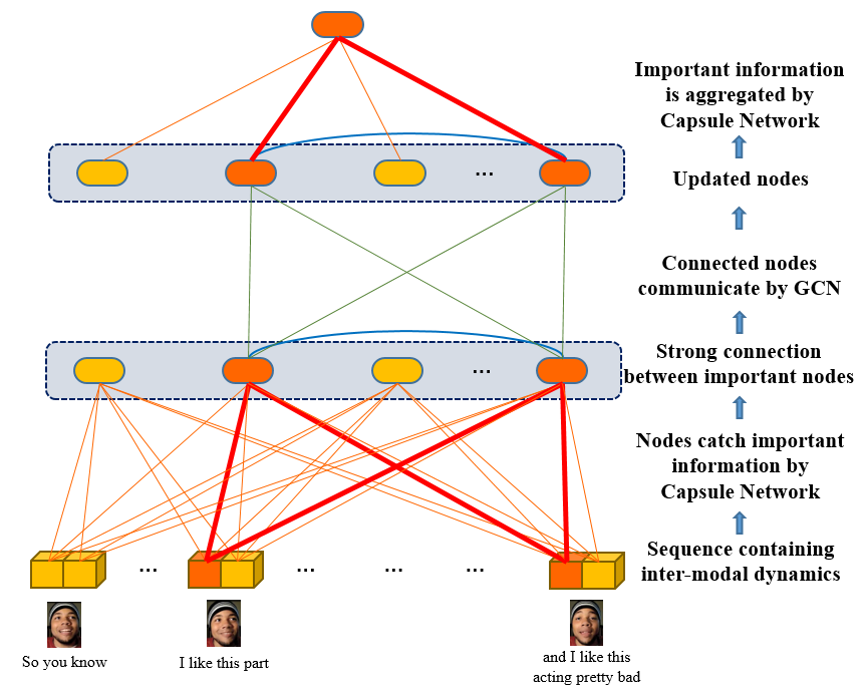}
    \caption{Illustration of the aggregation capability of the GraphCAGE. The color of the links depend on the values of routing coefficients. Red means large value and important information such as the word "like", whereas orange means common information. Blue links indicate edges between nodes. Note that our model can pay attention to critical information from different time steps although they are far from each other.}
    \label{fig:aggregation}
\end{figure}

Existing models commonly implement forced word-alignment before training\cite{Zadeh2018Memory,MOSEI,MFM,MCTN,Gu2018Multimodal,MFRM} to solve the problem of “unaligned” nature, which aligns the visual and acoustic features to the resolution of words before inputting them into model. However, such word-alignment \cite{P2FA} is time-consuming and not feasible because it requires detailed meta-information about the datasets. Moreover, it may lead to inadequate interactions between modalities as the interactions are not limited to the span of one word. Therefore, the issue of long-range dependency still exists. In addition, owing to heavy reliance on RNN, previous models are usually difficult to train and require plenty of time to infer. Recently, some transformer-based models\cite{MULT,2019Factorized,9053762} which can compute in parallel in the time dimension have been proposed to avoid problems of RNN and better explore long-range dependency. Nevertheless, they fail to obtain highly expressive and refined representation of sequences because transformer \cite{transformer} is a sequence model which cannot sufficiently fuse information from all time steps.

In this paper, we propose an end-to-end model called \textbf{\underline{Graph}} \textbf{\underline{C}}apsule \textbf{\underline{Ag}}gr\textbf{\underline{e}}gation (GraphCAGE) that can compute in parallel in the time dimension by converting unaligned multimodal sequential data into graphs and explicitly learn long-range dependency by the aggregation capability of Capsule Network and graph-base neural model. GraphCAGE consists of two stages: graph construction and graph aggregation. The former first implements modality fusion by cross-modal transformer, then applies Dynamic Routing of Capsule Network and self-attention to create nodes and edges, respectively. This module can significantly solve the problem of long-range dependency because the nodes can proportionally absorb information from every time step by routing mechanism. The latter stage combines Graph Convolutional Network (GCN) with Capsule Network to further aggregate information from nodes and finally produces high-level and refined representation of the graph. We illustrate the aggregation capability of our model in Figure~\ref{fig:aggregation}. Additionally, routing mechanism equips GraphCAGE with interpretability because we are able to observe the values of routing coefficients to figure out the contributions from different elements. We will discuss the interpretability in Section~\ref{sec:interpretability}.

In brief, the main contributions of this work are listed below:
\begin{itemize}
\item We propose a novel architecture called GraphCAGE to model unaligned multimodal sequences. GraphCAGE applies Dynamic Routing of Capsule Network to construct node, which enables the model to process longer sequence with stronger ability of learning long-range dependency. Taking advantage of aggregation capability of Capsule Network, GraphCAGE produces high-expressive representations of graphs without any loss of information.
\item With sequences transformed into graphs, GraphCAGE can model sequence without RNN, which prevents gradient vanishing or exploding during training. Moreover, computing in parallel greatly boosts efficiency in inferring time.
\item Applying Capsule network in node construction and graph aggregation, GraphCAGE is interpretable owing to routing mechanism. With larger routing coefficients indicating greater contribution, we can figure out what information our model focuses on to make predictions.
\item The proposed GraphCAGE model achieves state-of-the-art performance on two widely-used datasets. In addition, the extensive experiments in Section~\ref{sec:dependency} and Section~\ref{sec:interpretability} on routing coefficients demonstrate that our model explicitly explores long-range dependency with interpretation provided.\end{itemize}
\fancyhead{}

\section{Related Works}
\subsection{Human Multimodal Language Analysis}
Multimodal language learning aims at learning representations from multimodal sequences including textual, visual and acoustic modalities\cite{RMFN,MFM}. A lot of previous studies\cite{Modality-Trans,ghosal-etal-2018-contextual,Hai2018Seq,MOSEI,Zadeh2018Multi,Zadeh2018Memory} regard RNN such as LSTM and GRU as the default architecture for sequence modeling and they focus on exploring intra- and inter-modal dynamics for word-aligned multimodal sequences. For example, Zadeh et al. propose Memory Fusion Network which is constructed by LSTMs and gated memory network to explore view-specific and cross-view interactions\cite{Zadeh2018Memory}. In \cite{Zadeh2018Multi}, Multi-attention Recurrent Network is composed of LSTMs and multi-attention block in order to model both dynamics above. With RNN being the main modules, they are confronted with the problems of training and long inferring time. Recently, \cite{MULT,DBLP:journals/corr/abs-2011-13572,DBLP:journals/corr/abs-2010-11985} propose alternative networks to model unaligned multimodal sequences. Tsai et al.\cite{MULT} use cross-modal transformer and self-attention transformer to learn long-range dependency. However, the temporal information is collected by self-attention transformer which is a sequence model, implying that fusion among different time steps is not sufficient. In contrast, our proposed GraphCAGE replaces the self-attention transformer with graph-based model which produces more refined and high-level representations of sequences. In \cite{DBLP:journals/corr/abs-2011-13572} and \cite{DBLP:journals/corr/abs-2010-11985}, sequences are transformed into graphs and GCN is applied to learn long-range dependency, which not only avoid the problems of RNN but also successfully model unaligned multimodal sequences. Nevertheless, they implement graph pooling and edge pruning to drop some nodes in order to obtain the final representation of graph, leading to information loss. In contrast, GraphCAGE effectively retains all information with Capsule Network which applies Dynamic Routing instead of pooling to aggregate features.

 \begin{figure*}[t]
    \centering
    \includegraphics[width=0.8\textwidth]{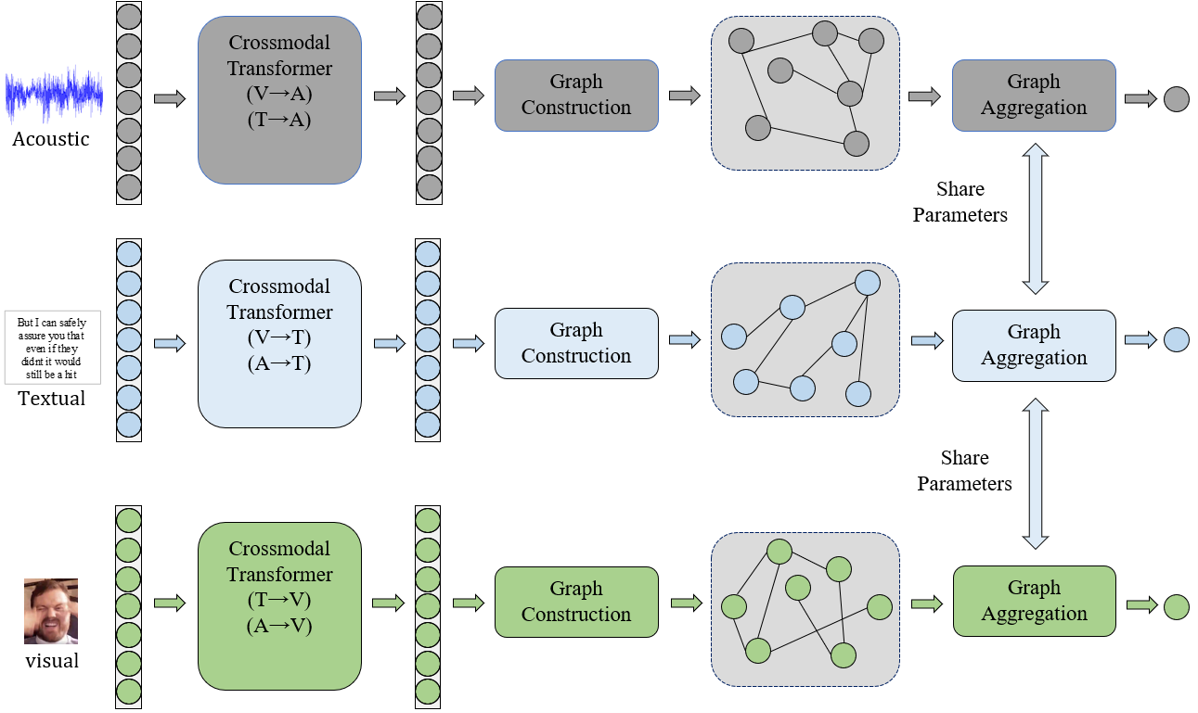}
    \caption{The Schematic Diagram of our proposed GraphCAGE.}
    \label{fig:algorithm}
\end{figure*}

\subsection{Capsule Network}
Capsule Network is first proposed in \cite{Capsule_Network} and is improved in \cite{e2018matrix} and \cite{kosiorek2019stacked}, which is designed for image features extraction. In general, Capsule Network can not only effectively fuse information from numerous elements into highly expressive representations without information loss, but also reveal the contributions from different elements to the representations by routing mechanism. In \cite{Capsule_Network}, the authors claim that pooling will destroy the robustness of the model because some valuable features are ignored by pooling layer. In order to retain these features, pooling layer is replaced with Dynamic Routing for the transmission of information between layers, bringing the benefit of no information loss. In \cite{MRM}, the proposed Multimodal Routing is designed based on Capsule Network and provides both local and global interpretation, verifying the fact that Dynamic Routing of Capsule Network can equip model with interpretability. Inspired by Dynamic Routing, our proposed GraphCAGE uses Capsule Network to construct node from features containing inter-modal dynamics. In addition, the final representations of graphs are also created by Capsule Network. As a result of efficient transmission of information and great aggregation capability of Capsule Network, our GraphCAGE can effectively learn long-range dependency and explicitly model unaligned multimodal sequences with interpretation ability provided and no information loss.

\subsection{Graph Neural Network}
As graph-structured data is widely used in many research fields, a series of Graph Neural Networks (GNN) have been introduced in recent years \cite{DIN,Micheli2009Neural,4700287,SEAL}. Among them, Graph Convolutional Network (GCN) \cite{GCN} is the most popular because of its superior performance on various tasks. Informed by the fact that GCN can effectively aggregate information of related nodes, we apply GCN to integrate related nodes which contain information from various time steps. By this way, the issue of long-range dependency is solved because even the information from two distant time steps can directly communicate with each other. In most cases, the final representation of a graph is obtained by graph pooling \cite{DBLP:journals/corr/abs-2011-13572,GraphSAGE,Diffpool}. Similarly, in order to obtain  high-level graph representation, edge pruning\cite{DBLP:journals/corr/abs-2010-11985} is usually applied in each GCN layer. However, pooling and pruning may rudely drop some important nodes, leading to the loss of information. As we conduct Dynamic Routing of Capsule Network instead of pooling or pruning after GCN, our proposed GraphCAGE model produces high-level and refined representations of sequences without the loss of information.

\section{Proposed Method}
In this section, we elaborate our proposed GraphCAGE with its diagram illustrated in Figure~\ref{fig:algorithm}. Our GraphCAGE consists of two stages including graph construction and graph aggregation. In the first stage, multimodal sequences are transformed into graphs with nodes and edges created by Capsule Network and self-attention respectively, which enables our model to compute in parallel in the time dimension. In the second stage, each graph is condensed into a representative vector via Graph Convolutional Network (GCN) and Capsule Network. Fundamentally, the Capsule Network in the first stage integrates information of every time step into each node, then the GCN and the Capsule Network in the second stage further aggregate information of nodes, which equips our model with excellent capability of learning long-range dependency.

\subsection{Graph Construction}
To construct a graph, we need to first create nodes from sequence, then define edges based on these created nodes. All the nodes and edges comprise the graph which contains sufficient information about sentiment and long-range dependency.

\begin{figure}[t]
    \centering
    \includegraphics[width=0.4\textwidth]{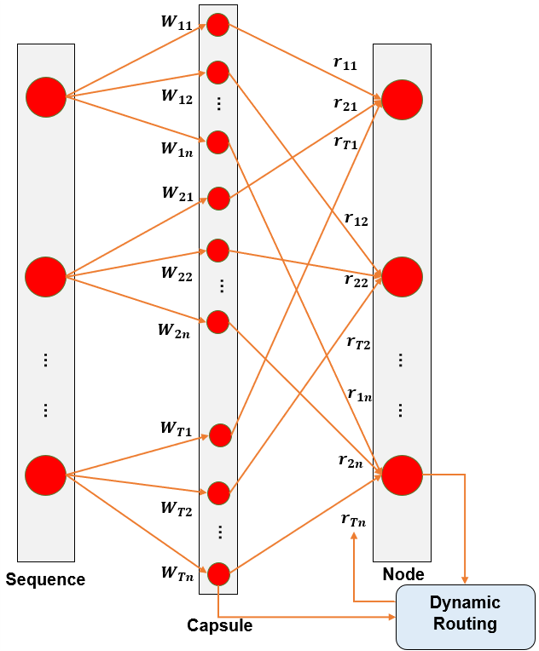}
    \caption{The Schematic Diagram of Node Definition. Due to conciseness, routing mechanism is only presented for $r_{Tn}$. In fact, every routing coefficient is updated at every iteration by routing mechanism.}
    \label{fig:CapsNet}
\end{figure}

\subsubsection{Node definition}
In order to create node containing information of interactions between different modalities, we first input features of textual, acoustic and visual modalities into cross-modal transformers\footnote{More detail about cross-modal transformer can be found in the link \url{https://github.com/kenford953/GraphCAGE}}\cite{MULT}:
\begin{equation}
\label{eu_eqn}
\begin{split}
Z^{t} &= CT^{v\rightarrow t}(X^{t},X^{v}) \oplus {CT^{a\rightarrow t}(X^{t},X^{a})}\cr
Z^{a} &= CT^{t\rightarrow a}(X^{a},X^{t}) \oplus {CT^{v\rightarrow a}(X^{a},X^{v})}\cr
Z^{v} &= CT^{t\rightarrow v}(X^{v},X^{t}) \oplus {CT^{a\rightarrow v}(X^{v},X^{a})}\cr
\end{split}
\end{equation}
where $X^{\{t,a,v\}}\in\mathbb{R}^{{d^{\{t,a,v\}}}\times{T^{\{t,a,v\}}}}$ denotes the inputted unimodal sequence with $d^{\{t,a,v\}}$ being the dimensionality of features and $T^{\{t,a,v\}}$ being the sequence length. $CT^{\alpha \rightarrow \beta}$ is the cross-modal transformer translating $\alpha$ modality into $\beta$ modality with $\oplus$ being the operation of concatenation. For conciseness, we denote $m\in{\{t,a,v\}}$ as a specific modality in the rest of this paper. The outputs $Z^{m}\in\mathbb{R}^{{d}\times{T^{m}}}$ contain  inter-modal dynamics but long-range dependency is still understudied because the output of cross-modal transformer is still a sequence which requires adequate fusion at the time dimension to explore the interactions between distant time steps. Capsule Network is an excellent model to figure out relations among various elements. Therefore, we apply Capsule Network to construct node from the output sequence $Z^{m}$ in order to properly fuse information from a large number of time steps. We illustrate the node definition in Figure~\ref{fig:CapsNet}. As shown in Figure~\ref{fig:CapsNet}, we first create capsules as:
\begin{equation} \label{capsule create}
Caps^{m}_{i,j}={W^{m}_{i,j}}{Z^{m}_{i}}
\end{equation}
where $Z^{m}_{i}\in\mathbb{R}^{d}$ denotes the features of the $i^{th}$ time step of sequence $Z^{m}$ with $W^{m}_{i,j}\in\mathbb{R}^{d_c\times{d}}$ being the trainable parameters. $Caps^{m}_{i,j}\in\mathbb{R}^{d_c}$ means the capsule from the $i^{th}$ time step and it is used for constructing the $j^{th}$ node. Then, we define nodes based on these capsules and Dynamic Routing as Algorithm~\ref{algorithm:construction} shows. Specifically, a node is defined by the weighted sum of corresponding capsules as shown below:
\begin{equation} \label{node definition}
N^{m}_{j}=\sum_{i}Caps^{m}_{i,j}\times{r^{m}_{i,j}}
\end{equation}
where $N^{m}_{j}$ denotes the embedding of the $j^{th}$ node and $r^{m}_{i,j}$ is the routing coefficient assigned to capsule $Caps^{m}_{i,j}$. It is worth noting that for a total of $p$ iterations, all routing coefficients are normalized by softmax and updated based on inner product between the embeddings of capsule and node in every iteration step. The equations for updating $r^{m}_{i,j}$ are shown as below:

\begin{equation} \label{routing softmax}
r^{m}_{i,j}=\frac{exp(b^{m}_{i,j})}{\sum_{j}exp(b^{m}_{i,j})}
\end{equation}

\begin{equation} \label{routing update}
b^{m}_{i,j}\leftarrow b^{m}_{i,j}+Caps^{m}_{i,j}\odot{N^{m}_j}
\end{equation}
where $b^{m}_{i,j}$ means the routing coefficient before normalization, which is initialized to zero before iteration begins. $\odot$ denotes the operation of inner product. By comparing the values of routing coefficients, we can understand how much information from a specific time step flows into a node, which provides interpretation. With Capsule Network applied to construct node, our model can effectively learn long-range dependency, because nodes contain information from the whole range of sequence and more informative time steps will be assigned larger routing coefficients.

\begin{algorithm}[t]
\caption{Node Definition By Dynamic Routing} %算法的名字
\label{algorithm:construction}
\begin{algorithmic}[1]
\Statex \textbf{Input:}%算法的输入,利用 \\ 进行换行
capsules $Caps^{m}_{i,j}$
\Statex \textbf{Output:} %算法的结果输出
nodes $N^{m}$
\State Initialize all routing coefficients to zero as $b^{m}_{i,j}=0$
\For{p iterations}
    \State Normalize all routing coefficients as Eq.~\ref{routing softmax}
    \State Create node $N^{m}_{j}$ as Eq.~\ref{node definition}
    \State Update all routing coefficients as Eq.~\ref{routing update}
\EndFor
\State \Return nodes $N^{m}$
\end{algorithmic}
\end{algorithm}

\subsubsection{Edge definition}
After node construction, edges are created by the self-attention mechanism over the nodes:
\begin{equation} \label{adjacent}
A^{m} = f(\frac{(W^{m}_q{N^{m}})^T(W^{m}_k{N^{m}})}{d_c})
\end{equation}
where $A^{m}\in\mathbb{R}^{n\times{n}}$ is the adjacency matrix and $N^{m}\in\mathbb{R}^{d_c\times{n}}$ denotes the overall node embeddings with $n$ being the number of nodes. $W^{m}_q,W^{m}_k\in\mathbb{R}^{d_c\times{d_c}}$ are learnable parameters. $f$ is the nonlinear activation function which is set to ReLU, and $T$ means the matrix transpose operation. With ReLU as our activation function, the negative links between nodes can be effectively filtered out \cite{DBLP:journals/corr/abs-2011-13572} (a negative link implies that a direct connection between these two nodes is not necessary).

It is worth noting that Capsule Network has a large number of trainable parameters. As a result, we apply L2 Regularization on these parameters to alleviate overfitting as:
\begin{equation} \label{eu_eqn}
L_{reg} = \lambda(\sum_{i}^{T^t}\sum_{j}^{n}{\Vert W^t_{i,j}\Vert}^2 + \sum_{i}^{T^a}\sum_{j}^{n}{\Vert W^a_{i,j}\Vert}^2 + \sum_{i}^{T^v}\sum_{j}^{n}{\Vert W^v_{i,j}\Vert}^2 )
\end{equation}
where $\lambda$ is a hyper-parameter which reflects the importance of the loss function. Therefore, during training, the total loss function is the Mean Absolute Error (MAE) plus $L_{reg}$.

As we finish constructing nodes and edges, a graph which contains rich inter-modal dynamics and reliable connections between related nodes has been created. Our graph construction method is informed by recent graph-based architectures for sequential data, but is distinct from all of them in the method of node construction. For example, in \cite{DBLP:journals/corr/abs-2011-13572} and \cite{DBLP:journals/corr/abs-2010-11985}, the authors define node and edge based on multimodal sequences processed only by Feed-Forward-Network, which causes that the created graph is not highly expressive because the node embedding is not built on high-level features. Moreover, they regard every time step as a node and only depend on GCN to learn long-range dependency, leading to insufficient learning. Contrary to them, our model first uses cross-modal transformer to obtain high-level features which contain inter-modal dynamics, then constructs node based on these features by Capsule Network which enables node to properly gain information from a great quantity of capsules. Note that the number of nodes here is significantly fewer than the length of the input sequence. By this way, each node is built on various time steps and the created graph is highly expressive and also is easier to be processed because of a small number of nodes. In the next stage, we illustrate how we conduct message passing between nodes and extract high-level representation from the graph.

\subsection{Graph Aggregation}
In most cases, representation of graph is extracted by GCN followed by graph pooling or edge pruning to dump some redundant nodes. However, it is hard to avoid dropping valuable nodes which causes the loss of information. To prevent this problem, we retain GCN due to its excellent capability of exchanging information among nodes, and replace pooling or pruning with Capsule Network to prevent information from being lost. %The graph aggregation is totally run for $r$ iterations.
The graph aggregation consists of inner-loop and outer-loop. The relationship between inner-loop and outer-loop can be explained in this way: in every iteration of outer-loop, all iterations of inner-loop will be performed. As for the proposed method, Graph convolution is performed in outer-loop and the Dynamic Routing is performed in the inner-loop. So, in every iteration of graph convolution, we will perform $p$ iterations of Dynamic Routing to obtain a graph representation. The equations for the $k^{th}$ iteration of graph convolution are shown below:
\begin{equation} \label{eu_eqn}
\begin{split}
n^{m,k} &=W^k N^{m,k-1} (A^{m}+I)\\
N^{m,k} &= f(W^{k}_{o} n^{m,k})
\end{split}
\end{equation}
where $N^{m,k}$ denotes the node embedding at the $k^{th}$ iteration and $N^{m,0}$ is the output node embedding in the graph construction stage ($1\leq k \leq 2$). $I$ denotes the identity matrix which is used to perform self-loop operation and $f$ is chosen to be the $tanh$ activation function. Note that $W^k$ and $W^{k}_{o}$ have no superscripts $m$ because we share all weights for three modalities in the graph aggregation stage. When all nodes are updated, %we concatenate and input them to Capsule Network to produce
we generate the final representation of the graph at the $k^{th}$ iteration using Capsule network. The Capsule network consists of $p$ iterations (i.e., the inner-loop iteration) to update the routing coefficients of the nodes. The equation is shown as below:
\begin{equation} \label{eu_eqn}
\begin{split}
%N^{\{t,a,v\}}_{k} &= \mathop{concat}^{n-1}_{j=0}(N^{\{t,a,v\}}_{j,k})\\
R^{m,k} = CapsNet(N^{m,k})
\end{split}
\end{equation}
where $R^{m,k}\in\mathbb{R}^{d_c}$ denotes the final representation at the $k^{th}$ iteration and the details of $CapsNet$ are shown in Algorithm~\ref{algorithm:aggregation}. Specifically, Dynamic Routing (i.e., the inner-loop)  contains normalization of routing coefficients, construction of representation and update of routing coefficients as shown below:
\begin{equation} \label{routing normalization}
r^{m,k}_{j}=\frac{exp(b^{m,k}_{j})}{\sum_{j}exp(b^{m,k}_{j})}
\end{equation}

\begin{equation} \label{representation definition}
R^{m,k}=\sum_{j}Caps^{m,k}_{j}\times{r^{m,k}_{j}}
\end{equation}

\begin{equation} \label{agg routing update}
b^{m,k}_{j}\leftarrow b^{m,k}_{j}+Caps^{m,k}_{j}\odot{R^{m,k}}
\end{equation}
where $Caps^{m,k}_{j}$ means the capsule created by the $j^{th}$ node at the $k^{th}$ graph convolution iteration (see Algorithm~\ref{algorithm:aggregation}). Note that different from the graph construction stage, each node only owns one capsule at each graph convolution iteration so only one subscript $j$ is enough for denoting the capsule.
\begin{algorithm}[t]
\caption{Capsule Network for Graph Aggregation} %算法的名字
\label{algorithm:aggregation}
\begin{algorithmic}[1]
\Statex \textbf{Input:} %算法的输入,利用 \\ 进行换行
 node embedding $N^{m,k}$
\Statex \textbf{Output:} %算法的结果输出
 representation $R^{m,k}$
\State Create capsules for each node as $Caps^{m,k}_{j}=W^{k}_{j} N^{m,k}_{j}$
\State Initialize all routing coefficients as $b^{m,k}_{j}=0$
\For{p iterations}
    \State Normalize all routing coefficients as Eq.~\ref{routing normalization}
    \State Create representation $R^{m,k}$ as Eq.~\ref{representation definition}
    \State Update all routing coefficients as Eq.~\ref{agg routing update}
\EndFor
\State \Return representation $R^{m,k}$
\end{algorithmic}
\end{algorithm}

\begin{table*}[t]
\centering
 \caption{ \label{t2}\textbf{Performance of GraphCAGE on two benchmark datasets.} The bold means the best performance. We put asterisk behind the result by our model which is not the best but close to SOTA(<1\%).}
 \label{tab:performance}
\resizebox{0.8\textwidth}{!}
{\begin{tabular}{c|ccccc|ccccc}
 \hline
 \multirow{2}{*}{Models} & \multicolumn{5}{c|}{CMU-MOSI} & \multicolumn{5}{c}{CMU-MOSEI}\cr
  & Acc7 & Acc2 & F1 & MAE & Corr & Acc7 & Acc2 & F1 & MAE & Corr\cr
 \hline
 \multicolumn{11}{c}{Recurrent Models}\cr
 \hline
 CTC+EF-LSTM & 32.2 & 73.7 & 73.5 & 1.038 & 0.594 & 41.7 & 65.3 & 76.0 & 0.799 & 0.625\cr
 CTC+LF-LSTM & 31.3 & 74.5 & 74.3 & 1.042 & 0.608 & 48.0 & 79.5 & 79.6 & 0.632 & 0.650\cr
 CTC+TFN & 32.4 & 77.9 & 75.0 & 1.040 & 0.616 & 49.3 & 79.5 & 78.9 & 0.613 & 0.673\cr
 CTC+MFN & 30.9 & 77.7 & 75.5 & 1.032 & 0.627 & 49.1 & 80.6 & 80.0 & 0.612 & \textbf{0.687}\cr
 \hline
 \multicolumn{11}{c}{Parallel Computing Models}\cr
 \hline
 MulT & 35.3 & 80.6 & 79.3 & 0.972 & 0.681 & 49.0 & 80.1 & 80.9 & 0.630 & 0.664\cr
 Multimodal Graph & 32.1 & 80.6 & 80.5 & \textbf{0.933} & 0.684 & \textbf{49.7} & 81.4 & 81.7 & \textbf{0.608} & 0.675\cr
 MTAG & 31.9 & 80.5 & 80.4 & 0.941 & \textbf{0.692} & 48.2& 79.1 & 75.9 & 0.645 & 0.614\cr
 GraphCAGE & \textbf{35.4} & \textbf{82.1} & \textbf{82.1} & \textbf{0.933} & 0.684$^*$ & 48.9$^*$ & \textbf{81.7} & \textbf{81.8} & 0.609$^*$ & 0.670\cr
 \hline
 \end{tabular}}
 \vspace{-0.3cm}
\end{table*}

As stated above, graph convolution enables related nodes communicate with each other and update node embedding, which helps our model further learn long-range dependency because nodes contain information from related time steps. Moreover, intra-modal dynamics are explored effectively because nodes are from two identical modalities. Finishing updating the nodes, Capsule Network is applied to aggregate all the nodes into a highly expressive representation with complete information transmission. More importantly, the highly expressive representation proportionally absorbs information from all nodes by Dynamic Routing, where larger routing coefficient will be assigned if information of the node is more valuable. By this way, interpretation is provided, indicating which node contributes most to the final representation. In contrast, many graph-based architectures roughly drop nodes by pooling or pruning to obtain the final representation, leading to the loss of information. In addition, interpretation of their models depends on the edges between related nodes, which reflect relations among different elements. But the contribution to prediction is not interpretable.

As intra- and inter-modal dynamics are effectively explored and long-range dependency is explicitly learned, we concatenate the graph representations of all the modalities at each iteration $k$ and apply fully-connected layers to predict sentiment labels.

\section{Experiments}
In this section, we evaluate our proposed model GraphCAGE on two frequently-used datasets: CMU-MOSI\cite{Zadeh2016MOSI} and CMU-MOSEI\cite{MOSEI}. We first show details about the datasets, baseline models, experimental settings, and then present the results with comparison among GraphCAGE and other baseline models. The remaining part of this section are illustrations about long-range dependency and interpretability.
\begin{comment}
\renewcommand{\arraystretch}{1.5} %控制行高
\begin{table}[h]
  \centering
  \fontsize{7}{8}\selectfont
  \caption{Details about extracted features of different datasets.}
  \label{tab:dataset details}
  \begin{tabular}{c|ccc|ccc}
  \hline
    \multirow{2}{*}{Dataset}&
    \multicolumn{3}{c|}{Embedding}&
    \multicolumn{3}{c}{Sequence Length}\cr
    &Textual&Acoustic&Visual&Textual&Acoustic&Visual\cr
    \hline
    CMU-MOSI& 300 & 5 & 20 & 50 & 375 & 500\cr
    CMU-MOSEI& 300 & 74 & 35 & 50 & 500 & 500\cr
    \hline
    \end{tabular}
\end{table}
\end{comment}

\renewcommand{\arraystretch}{1.5} %控制行高
\begin{table}[t]
\centering
\caption{Comparison with RNN-based models about inferring time on CMU-MOSI test set. Note that the inferring time is calculated based on the whole test set, which contains 686 video clips. the batch size and the environment are the same for all models.}
\label{tab:inferring time}
 {\begin{tabular}{c|c}
  \hline
  Models & Inferring Time (s)\cr
  \hline
  CTC+LF-LSTM & 8.926\cr
  CTC+TFN & 6.146\cr
  CTC+RAVEN & 19.369\cr
  \hline
  GraphCAGE & \textbf{3.813}\cr
  \hline
 \end{tabular}}
\end{table}

\renewcommand{\arraystretch}{1.5} %控制行高
\begin{table}[t]
  \centering
  \fontsize{7}{8}\selectfont
  \caption{An ablation study on the benefit of GraphCAGE's Capsule Network using unaligned CMU-MOSI.}
  \label{tab:ablation}
    \begin{tabular}{c|cc}
    \hline
    Models & Acc2 & F1 \cr
    \hline
    Graph Construction without Capsule Network & 76.1 & 76.7\cr
    Graph Aggregation with GAT & 75.9 & 77.0\cr
    Graph Aggregation with mean pooling & 77.0 & 77.0\cr
    Graph Aggregation with LSTM & 79.0 & 79.2 \cr
    \hline
    GraphCAGE & \textbf{82.1} & \textbf{82.1}\cr
    \hline
    \end{tabular}
\end{table}

\subsection{Datasets}
CMU-MOSI is a popular dataset for multimodal sentiment analysis which contains 2199 video clips. Each video clip is labeled with a real number within [-3, +3] which reflects the sentiment intensity, where +3 means strongly positive sentiment and -3 means strongly negative sentiment. In accordance with most prior works, various metrics are reported including 7-class classification accuracy ($Acc_7$), binary classification accuracy ($Acc_2$), Mean Absolute Error (MAE) , F1 score and the correlation of the model's prediction with humans. The total numbers of video clips for training set, validation set and testing set are 1284, 229 and 686, respectively.

CMU-MOSEI consists of 22856 video clips and we use 16326, 1871 and 4659 segments as training, validation and testing set. The reported metrics and sentiment label are the same as those of CMU-MOSI.

\subsection{Baseline Models}
We separate baseline models into two groups including recurrent models and parallel computing models.

Recurrent models include Early Fusion LSTM (EF-LSTM), Late Fusion LSTM (LF-LSTM), Tensor Fusion Network (TFN)\cite{Zadeh2017Tensor} and Memory Fusion Network (MFN)\cite{Zadeh2018Memory}. EF-LSTM and LF-LSTM simply concatenate features at input and output level, which apply LSTM\cite{17} to extract features and infer prediction. As stated in \cite{Zadeh2017Tensor}, these approaches fail to explore intra- and inter-modal dynamics due to simple concatenation. TFN effectively explores both dynamics with outer product adopted to learn joint representation of three modalities. MFN depends on systems of LSTM to learn interactions among modalities. However, EF-LSTM and MFN are word-level fusion methods which study aligned multimodal sequences and thus we combine connectionist temporal classification (CTC)\cite{CTC} with them to process the unaligned sequences. The CTC module we use comprises two components: alignment predictor and the CTC loss. The alignment predictor is chosen as a recurrent  networks. We train the alignment predictor while minimizing the CTC loss. Then, we multiply the  probability outputs from the alignment predictor to source  signals. The recurrent natures of the above models bring about some disadvantages including gradient vanishing or exploding, long inferring time and insufficient learning for long-time dependency.

Parallel computing models include Multimodal Transformer (MulT)\cite{MULT}, Multimodal Graph\cite{DBLP:journals/corr/abs-2011-13572} and Modal-Temporal Attention Graph (MTAG)\cite{DBLP:journals/corr/abs-2010-11985}, which disuse RNN to better explore long-range dependency within multimodal sequences. MulT extends Transformer network\cite{transformer} to model unaligned multimodal sequences by cross-modal transformer. Nevertheless, it utilizes self-attention transformer to integrate information from different time steps, which causes inadequate fusion at the time dimension because self-attention transformer is a sequence-to-sequence model and cannot fuse sequences at the time dimension. Multimodal Graph and MTAG both creatively adapt GCN to explore long-range dependency with problems of RNN avoided. However, they are confronted with information loss because of the operations of pooling and pruning.

\subsection{Experimental details}
Our model is developed on Pytorch and we choose Mean Absolute Error (MAE) as loss function for sentiment prediction task on CMU-MOSI and CMU-MOSEI datasets. Note that the total loss during training is MAE plus L2 Regularization loss. The optimizer is RMSprop and all hyper-parameters are selected by grid search. The textual, acoustic and visual features are extracted by GloVe word embedding\cite{pennington2014glove}, COVAREP\cite{Degottex2014COVAREP} and Facet\cite{FACET} respectively, with more details in \url{https://github.com/A2Zadeh/CMU-MultimodalSDK}. We specify the hyper-parameters and the features in our github link\footnote{The code for our model and details about hyper-parameters and features can be found in \url{https://github.com/kenford953/GraphCAGE}}. %The sequence length and the dimensionality of the extracted features vary with different datasets, which are shown in Table~\ref{tab:dataset details}.

\subsection{Results and Discussions}
The overall results are shown in Table~\ref{tab:performance} which indicates that our model outperforms both recurrent and parallel computing models on most of the metrics for two popular datasets. In general, based on the results that parallel computing models achieve better performance than recurrent models, we can infer that it is practical to apply model without recurrent structure to multimodal sequence modeling.

Comparing with recurrent models, GraphCAGE outperforms them by a considerable margin which implies that our model processes sequential data better than canonical RNN. Low performance on unaligned sequences by RNN-based models verifies the incompetence of recurrent network to model excessively long sequence which requires strong capability of learning long-range dependency. With aggregation capability of Capsule Network and the graph-based structure, GraphCAGE can effectively link remote but related time steps, which contributes to the explicit exploration of long-range dependency. Moreover, as Table~\ref{tab:inferring time} shows, the inferring time of our model is significantly reduced, demonstrating the high efficiency of our model which can compute in parallel in the time dimension.

As for parallel computing models, GraphCAGE achieves the best performance due to substantially longer memory and efficient transmission of information. Specifically, because GCN and Capsule Network in graph aggregation stage can realize more sufficient fusion at the time dimension than self-attention transformer, GraphCAGE explicitly explores long-range dependency and outperforms MulT. In addition, with Capsule Network applied to transmit information, our model achieves better performance than MTAG and Multimodal Graph which have shortcoming about the loss of information.

\subsubsection{Ablation Study} In order to verify the effectiveness of our graph construction and graph aggregation stages, we conduct ablation study on CMU-MOSI dataset as Table~\ref{tab:ablation} shows. Generally, the absence of Capsule Network in both stages of our model leads to drastic decline on performance, which indicates that they are critical to improve the ability of learning long-range dependency and enable our GraphCAGE to better model unaligned multimodal sequences.

 \begin{figure}[t]
    \centering
    \includegraphics[width=0.4\textwidth]{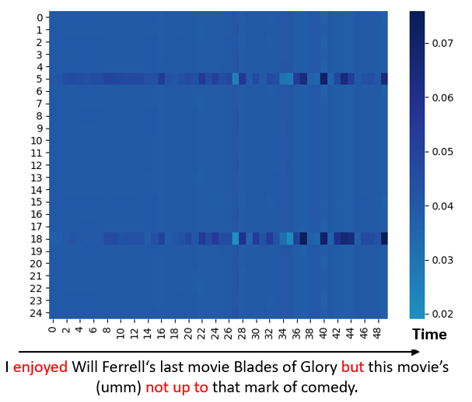}
    \caption{Visualization for routing coefficients of Capsule Network in graph construction stage. The element of the $i^{th}$ column and the $j^{th}$ row represents the value of routing coefficient $r_{ij}$ which is the same as $r_{ij}$ in Figure~\ref{fig:CapsNet}. Note that the values in each column are the routing coefficients from each time step. The presented routing coefficients are from textual modality for the sake of clear interpretation.}
    \label{fig:dependency}
\end{figure}

For model without Capsule Network in graph construction, we directly define each node embedding as the feature of each time step and edges are constructed by self-attention. Apparently, each node only contains information from one time step, which causes insufficient learning for long-range dependency. Moreover, owing to the long sequence length, the number of nodes is  excessively large. As a result, the latter GCN and Capsule Network are hard to  figure out the relations among these nodes. In contrast, our model first condenses information from sequence into a moderate number of nodes by Capsule Network, then models their relations by later layers, which improves the capability of linking remote but related time steps.

 \begin{figure*}[t]
    \centering
    \includegraphics[width=0.8\textwidth]{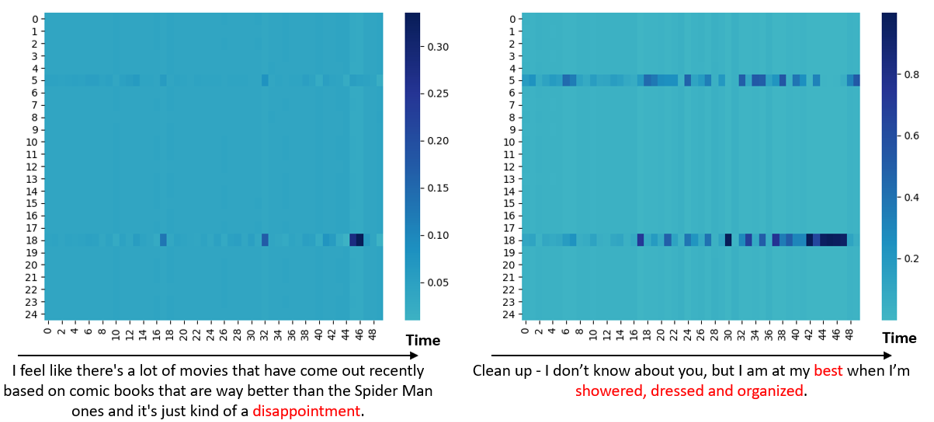}
    \caption{Visualization for routing coefficients of Capsule Network in Graph Construction stage. Similarly, The element represents the value of a routing coefficient and all routing coefficients come from textual modality. The left is a negative example whereas the right is a positive one.}
    \label{fig:interpretability}
\end{figure*}

For graph aggregation without Capsule Network, we retain the GCN part and design three aggregation methods including Graph Attention Network (GAT)\cite{GAT}, mean pooling and LSTM to replace the Capsule Network. Note that GAT applies attention mechanism to aggregate nodes and achieves excellent performance on various node classification tasks. However, based on the lower performance, we argue that GAT is not suitable for our model because we need to decode the nodes to predict a label rather than classify them. As for mean pooling, the final representation is the average of the embeddings of all nodes. Obviously, mean pooling is too simple to obtain highly expressive graph representation and it will cause the loss of information. For LSTM, it is slightly better than mean pooling because of more learnable parameters. However, the final representation is the last element of the output sequence. As a result, the input order may heavily affect the performance and we cannot figure out the best order because information of a node changes dynamically. In conclusion, applying Capsule Network to aggregate information of nodes is more suitable than other frequently-used aggregation methods because the final representation is refined by absorbing more important information by Dynamic Routing.

\subsubsection{Discussion of Long-range Dependency}
\label{sec:dependency} As we stated above, because of the adaptation of Capsule Network, GraphCAGE is skilled at modeling long sequences which requires excellent capability of learning long-range dependency. To present this ability in detail, as shown in Figure~\ref{fig:dependency}, we find an example of CMU-MOSEI and observe its routing coefficients in graph construction stage which reflect how much the model pays attention to specific information. Specifically, the sentiment of this example is obviously negative because of the word "but" and the phrase "not up to" in the last part of the sentence. However, some models with weak ability of learning long-range dependency may predict positive for this example based on the word "enjoyed" in the front part of the sentence. In contrast, our model attends to both parts of the sentence and pays more attention to the last part with larger routing coefficients assigned. Moreover, we found that the information prefers to flow into the fifth and eighteenth nodes which communicate by GCN and are integrated by Capsule Network later. Presumably it is because the distance between these two nodes is moderate which prevents our model from overly focusing on specific part of the sequence and the later GCN and Capsule Network enable our model to figure out the relations among important parts of the sentence. So we believe that even if the exact sentiment requires contextual information, our model can correctly predict sentiment with excellent capability of connecting remote but related elements.

\subsubsection{Discussion of Interpretability} \label{sec:interpretability}Interpretation helps us to figure out how the model comes to a prediction from a large number of time steps, which is useful for improving performance on different datasets. To provide interpretation, we adapt Capsule Network into our model where the routing coefficients reflect how much information from the corresponding time step flows into the next layer. As shown in Figure~\ref{fig:interpretability}, we observe the values of routing coefficients from textual modality of two examples with different sentiments. For the left example, information of the word "disappointment" is highlighted by the largest routing coefficient, indicating our model predicts the negative sentiment mostly depending on it. As for the right example, our model successfully catches the important positive words "best", "showered", "dressed" and "organized" by assigning larger routing coefficients to them. Based on the analysis above, we can safely draw a conclusion that GraphCAGE actually understands which element leads to specific sentiment and it provides interpretation for us to find out what information contributes to the prediction.

\section{Conclusion}
In this paper, we develop a model called GraphCAGE for multimodal sentiment analysis using unaligned multimodal sequences which are too long to model by recurrent networks. For the purpose of explicitly learning long-range dependency, our model adapts Capsule Network to transmit information and applies Graph Convolutional Network to explore relations among different time steps, which can avoid the loss of information and contributes to the interpretation. Moreover, modeling sequences with graph-based structure instead of RNN prevents various problems like gradient vanishing or exploding. Extensive experiments with routing coefficients verify the effectiveness of the adaptation of Capsule Network and GCN. Experiments on two popular datasets show that GraphCAGE achieves SOTA performance on modeling unaligned multimodal sequences.

\begin{acks}
This work was supported by the National Natural Science Foundation of China under Grant 62076262.
\end{acks}

\clearpage
\bibliographystyle{ACM-Reference-Format}
\bibliography{main}

\end{document}